# Clinical Domain Knowledge-Derived Template Improves Post Hoc AI Explanations in Pneumothorax Classification

Han Yuan[1†], Chuan Hong[2†], Pengtao Jiang[3], Gangming Zhao[4], Nguyen Tuan Anh Tran[5], Xinxing Xu[6], Yet Yen Yan[7], Nan Liu[1,8,9✉]

[1] Centre for Quantitative Medicine, Duke-NUS Medical School, Singapore

[2] Department of Biostatistics and Bioinformatics, Duke University, USA

[3] College of Computer Science, Nankai University, China

[4] Faculty of Engineering, The University of Hong Kong, China

[5] Department of Diagnostic Radiology, Singapore General Hospital, Singapore

[6] Institute of High Performance Computing, Agency for Science, Technology and Research, Singapore

[7] Department of Radiology, Changi General Hospital, Singapore

[8] Programme in Health Services and Systems Research, Duke-NUS Medical School, Singapore

[9] Institute of Data Science, National University of Singapore, Singapore

[†] Equal contribution

✉ Correspondence: Nan Liu, Centre for Quantitative Medicine, Duke-NUS Medical School, 8 College Road, Singapore 169857, Singapore. Phone: +65 6601 6503. Email: liu.nan@duke-nus.edu.sg

## Abstract

Background: Pneumothorax is an acute thoracic disease caused by abnormal air collection between the lungs and chest wall. Recently, artificial intelligence (AI), especially deep learning (DL), has been increasingly employed for automating the diagnostic process of pneumothorax. To address the opaqueness often associated with DL models, explainable artificial intelligence (XAI) methods have been introduced to outline regions related to pneumothorax diagnoses made by DL models. However, these explanations sometimes diverge from actual lesion areas, highlighting the need for further improvement.

Method: We propose a template-guided approach to incorporate the clinical knowledge of pneumothorax into model explanations generated by XAI methods, thereby enhancing the quality of these explanations. Utilizing one lesion delineation created by radiologists, our approach first generates a template that represents potential areas of pneumothorax occurrence. This template is then superimposed on model explanations to filter out extraneous explanations that fall outside the template's boundaries. To validate its efficacy, we carried out a comparative analysis of three XAI methods (Saliency Map, Grad-CAM, and Integrated Gradients) with and without our template guidance when explaining two DL models (VGG-19 and ResNet-50) in two real-world datasets (SIIM-ACR and ChestX-Det).

Results: The proposed approach consistently improved baseline XAI methods across twelve benchmark scenarios built on three XAI methods, two DL models, and two datasets. The average incremental percentages, calculated by the performance improvements over the baseline performance, were 97.8% in Intersection over Union (IoU) and 94.1% in Dice Similarity Coefficient (DSC) when comparing model explanations and ground-truth lesion areas. We further visualized baseline and template-guided model explanations on radiographs to showcase the performance of our approach.

Conclusions: In the context of pneumothorax diagnoses, we proposed a template-guided approach for improving AI explanations. This approach not only aligns model explanations more closely with clinical insights but also exhibits expandability to other thoracic diseases. We anticipate that our template guidance will forge a fresh approach to elucidating AI models by integrating clinical domain expertise.

Keywords: Pneumothorax Diagnosis, Convolutional Neural Networks, Explainable Artificial Intelligence, Saliency Map, Grad-CAM, Integrated Gradients

# 1 Introduction

Pneumothorax is an acute thoracic disease caused by abnormal air collection in the pleural space between the lungs and chest wall [1]. Timely intervention is crucial to prevent pneumothorax from evolving into a life-threatening emergency [2]. In clinical practice, pneumothorax is usually diagnosed by radiologists on a chest radiograph - a process that demands considerable expertise and expert efforts. Recent advancements suggest that this process can be automated using artificial intelligence (AI), especially deep learning (DL) models such as convolutional neural networks (CNNs). For instance, EfficientNet B3 [3] has demonstrated high accuracy in classifying pneumothorax of various sizes, with the area under the receiver operating characteristic curve (AUROC) ranging from 88% to 96% [2]. Xception [4] further advanced the classification capability, achieving an AUROC of 99% on an open-access dataset [5]. While these DL-based classifiers have exhibited high-fidelity classification ability, their complexity poses a challenge: Comprising numerous interconnected neurons with intricate relationships, their decision-making processes are often opaque and challenging to interpret [6]. This complexity can hinder radiologists' acceptance and trust in these AI tools, thereby affecting their practical application in real-world settings [7,8].

To solve this problem, researchers have introduced various explainable artificial intelligence (XAI) methods to chest radiograph analysis. For instance, Mosquera et al. [9] applied class activation maps (CAM) [10] to identify regions in chest radiographs that significantly influence the disease diagnosis. Feng et al. [11] and Wang et al. [4] used Grad-CAM [12], a variant of CAM, to pinpoint the specific pixels on chest radiographs that contributed most to model predictions. These heatmaps partially alleviate radiologists' concerns regarding the trustworthiness of DL models [6]. However, a recent benchmarking study pointed out a notable result: even with a high-accuracy DenseNet-121 [13] achieving an AUROC of 99.3% in the pneumothorax classification, the areas highlighted by the model only coincided with 7.0% of the actual lesion areas as delineated by radiologists [14]. Similarly, Rocha et al. developed a ResNet-50 [15] with an AUROC of 85.4% in classifying pneumothorax, yet its explanations attained an Intersection over Union (IoU) of 17.6% when assessed using lesion areas delineated by coarse bounding boxes [16]. Giachanou et al. reported IoUs ranging from 3.1% to 15.1% across a variety of model explanations for pneumothorax diagnoses [17]. These identified discrepancies between classification and explanation capabilities underline the urgent need to improve existing model explanations [14].

Leveraging prior clinical knowledge is one promising direction for enhancing model explanations. Specifically, pneumothorax occurs in the pleural space between the lungs and chest walls [1]. This clinically validated information could serve as invaluable prior knowledge to improve model explanations [18]. Previous studies have successfully utilized disease location information in pneumothorax classification and localization. Crosby et al. [19] capitalized on the observation that pneumothorax typically occurs in apex areas of chest radiographs. Therefore, they segmented the upper third of chest radiographs for pneumothorax classification, achieving enhanced accuracy. However, it remains unclear whether model explanations can also take advantage of the location information. Addressing this, Jung et al. [7] identified common thoracic disease patterns on chest radiographs, directing models to focus on typical disease locales, which in turn enhanced both classification and explanation quality. However, their method requires an exhaustive labeling of eight common thoracic diseases and is inappropriate for resource-limited settings where only diagnostic labels of a single disease are available.

To overcome aforementioned limitations, we propose a template-guided approach that crafts a template covering potential occurrence areas of pneumothorax to guide model explanations generated by baseline XAI methods. We illustrate the performance of our approach through comparative experiments of three XAI methods with and without our template guidance. We hope our template-guided approach provides a novel perspective for incorporating clinical knowledge into the explanation of other thoracic conditions.

## 2 Methods

AI models, especially CNNs, have become the mainstream backbones for chest radiograph classification, with various XAI methods accompanied to interpret their diagnostic processes [12,20,21]. Despite these advancements, a recent study [14] indicates that model explanations provided by the pneumothorax classifier fail to match ground truth lesion areas, suggesting a need for further improvement. To bridge this gap, we propose a template-guided approach for existing XAI methods in the context of pneumothorax diagnoses. This section outlines our methodology, starting with an introduction of notations followed by a detailed description of CNNs' training strategy. We then illustrate three well-established explanation methods for CNNs. The section concludes with our proposed approach that guides model explanations with a clinical knowledge-derived template.

### 2.1 Notations

We first introduce key notations for subsequent illustrations of classifier training and explanation. For the pneumothorax classification task, we denote the nonoverlapping training, validation, and test dataset as $D^{train}$, $D^{val}$, and $D^{test}$, respectively. Each dataset consists of pairs of images and corresponding image-level binary labels, structured identically. As an illustrative example, we consider the training dataset $D^{train}$, which includes $N_{train}$ samples:

$$D_{train} = \{(I_i^{train}, Y_i^{train}), i = 1,2, \dots, N_{train}\}. \tag{1}$$

$I_i^{train}$ designates a two-dimensional image with a width of $W_0$ and a height of $H_0$. $Y_i^{train} \in \{0,1\}$ is the ground truth label by radiologists and $Y_i^{train} = 1$ states that $I_i^{train}$ is diagnosed with pneumothorax. $I_i^{train}$ consists of $W_0 \times H_0$ pixels and $p_{w,h}(I_i^{train})$ denotes a pixel in $I_i^{train}$ whose coordinate of width and height is $(w, h)$:

$$I_i^{train} = \{p_{w,h}(I_i^{train}), w = 1,2, \dots, W_0, h = 1,2, \dots, H_0\}. \tag{2}$$

Each $p_{w,h}(I_i^{train})$ comprises three elements $e_{w,h,c}(I_i^{train})$ standing for pixel values in channel $c$ of red, green, or blue:

$$p_{w,h}(I_i^{train}) = \{e_{w,h,c}(I_i^{train}), c = red, green, blue\}. \tag{3}$$

Based on the input of $e_{w,h,c}(I_i^{train})$ and the output target of $Y_i^{train}$, the pneumothorax classifier is trained and subsequently explained. Model explanations are typically generated by initially calculating the importance of pixels and then shortlisting pixels with importance values exceeding a pre-determined threshold to constitute the important sub-region [14,22]. Our template-guided approach relies on a radiologists-annotated lesion delineation $A_T^{train}$ of a single image $I_T^{train}$ from $D_{train}$. Additionally, for assessing the alignment between model explanations and real lesion areas, pneumothorax samples $I_i^{test}$ in test dataset $D^{test}$ are also annotated with pixel-level lesion areas $A_i^{test}$. $A_T^{train}$ holds the same structure as $A_i^{test}$ and we use $A_i^{test}$ as an illustrative instance. $A_i^{test}$ is a two-dimensional image and consists of $W_0 \times H_0$ elements $a_{w,h}(A_i^{test}) \in \{0,1\}$. Decided by radiologists, $a_{w,h}(A_i^{test}) = 1$ denotes that the pixel with the coordinate of $(w, h)$ in $I_i^{test}$ belongs to the lesion areas:

$$A_i^{test} = \{a_{w,h}(A_i^{test}), w = 1,2, \dots, W_0, h = 1,2, \dots, H_0\}. \tag{4}$$

It is important to note that lesion annotations $A_T^{train}$ and $A_i^{test}$ are exclusively employed for model explanations. The model training of pneumothorax classifiers follows the standard paradigm that uses binary diagnostic labels $Y_i^{train}$ and $Y_i^{val}$ [23].

## 2.2 Image classifier training

CNNs have achieved outstanding performance in various thoracic disease classification tasks [24]. In general, the image classifier training is to find a set of parameters that minimizes the difference between CNNs' predictions and ground truth labels in the training set. Formally, with the training dataset $D^{train}$, we aim to optimize a model $f_\theta$ parameterized by $\theta$. The model takes input $e_{w,h,c}(I_i^{train})$, $f_\theta$ produces an output $f_\theta(I_i^{train})$. The optimization objective is to minimize the difference $d$ between $f_\theta(I_i^{train})$ and sample labels $Y_i^{train}$ for all samples in $D^{train}$. The cumulative difference over all training samples known as loss function $l(\theta; D^{train})$ is expressed as:

$$l(\theta; D^{train}) = \frac{1}{N_{train}} \sum_i d(f_\theta(I_i^{train}), Y_i^{train}).$$  (5)

To avoid overfitting of $f_\theta$, the validation dataset $D^{val}$ is applied to early stop the optimization procedure. If the loss $l(\theta; D^{val})$ has not decreased for a pre-defined epoch number $N_{epoch}$, the iteration of $\theta$ will be terminated. The last $\theta$ that led to a decrease in $l(\theta; D^{val})$ is retained as the optimal parameter $\theta^*$:

$$\theta^* = argmin_\theta \left( l(\theta; D^{train}) \Big| l(\theta; D^{val}) \right).$$  (6)

After the determination of $\theta^*$, we measure the classification performance $M$ on the unseen test dataset $D^{test}$. An evaluation metric $m$ is used to assess the model performance by comparing the model prediction $f_{\theta^*}(I_i^{test})$ and the true label $Y_i^{test}$:

$$M(\theta^*; D^{test}) = \frac{1}{N_{test}} \sum_i m(f_{\theta^*}(I_i^{test}), Y_i^{test}).$$  (7)

## 2.3 Image classifier explanation

The developed model $f_{\theta^*}$ classifies an unseen image $I_i^{test}$ from the test dataset $D^{test}$ as $f_{\theta^*}(I_i^{test})$. We aim to further explain $f_{\theta^*}(I_i^{test})$ to both uncover the model decision logic and evaluate its trustworthiness [25,26]. A commonly used explanation paradigm calculates each pixel's importance $E(p_{w,h}(I_i^{test}))$ to the prediction $f_{\theta^*}(I_i^{test})$, and further identifies the focus area $R(I_i^{test})$ consisting of the most discriminative pixels towards model outputs [27]. Explanations are considered reliable if focus areas precisely match disease lesion areas annotated by human experts [14]. Within the explanation paradigm using focus areas, we introduce three mainstream XAI techniques of Saliency Map [20], Grad-CAM [12], and Integrated Gradients [21] to generate and evaluate model explanations. Concrete technical details of these techniques have been elaborated in their respective original publications [12,20,21]. Here we provide a concise overview to facilitate the downstream illustration of our template-guided approach.

As a pioneer in the image classifier explanation, Saliency Map [20] calculates the importance of $p_{w,h}(I_i^{test})$ through its forthright gradient towards $f_{\theta^*}(I_i^{test})$. Specifically, it computes $f_{\theta^*}(I_i^{test})$'s gradients with respect to every element $e_{w,h,c}(I_i^{test})$ in pixel $p_{w,h}(I_i^{test})$ and derives the pixel importance $E(p_{w,h}(I_i^{test}))$ as the largest absolute gradient among all channels.

Saliency Map depicts the impact of each pixel towards final model outputs while possibly outlines all recognizable objects in $I_i^{test}$ and fails to spotlight $R(I_i^{test})$ towards $f_{\theta^*}(I_i^{test})$ [10]. Grad-CAM [12] conjectures that the problem can be resolved by initially computing the pixel importance $E(p_{w,h}(I_{i,conv}^{test}))$ on the last convolutional layer $I_{i,conv}^{test}$, and subsequently reconstructing $E(p_{w,h}(I_i^{test}))$ through the bilinear interpolation of $E(p_{w,h}(I_{i,conv}^{test}))$.

Both Saliency Map and Grad-CAM depict the local changes in $f_{\theta^*}(I_i^{test})$ with respect to a small range of pixel values. However, if a pixel's possible values within a narrow range are always important towards $f_{\theta^*}(I_i^{test})$, the gradient saturates to zero, indicating the opposite conclusion that the pixel is trivial [21]. Integrated Gradients [21] solves this problem via computing the gradients sum of $m$ pseudo images interpolated between $I_i^{test}$ and a reference image $I_{ref}$ obtained by fusing all training images. Same as the previous two methods, Integrated Gradients output the pixel importance $E(p_{w,h}(I_i^{test}))$.

After obtaining $E(p_{w,h}(I_i^{test}))$ by different methods, a binarization cutoff value $v^*$ is used to outline the most important pixels and constitute the model focus region $R_{v^*}(I_i^{test})$. Explanations are considered reliable when $R_{v^*}(I_i^{test})$ highly overlaps with lesion areas $A_i^{test}$ [14]. Different metrics, e.g. $IoU$, are applied to quantify the performance $Q$ of model explanations on the test dataset $D^{test}$:

$$Q = \sum_i IoU(R_{v^*}(I_i^{test}), A_i^{test}).$$ (8)

## 2.4 Proposed template-guided explanation

As illustrated above, baseline XAI methods outline the important region $R_{v^*}(I_i^{test})$ from the whole area of $I_i^{test}$. However, domain knowledge elucidates that pneumothorax typically occurs in the pleural space between the lungs and chest walls [7,19]. Particularly, on an upright frontal radiograph, pneumothorax is recognized by non-dependent lucency that parallels the chest wall and displaces the visceral pleural line medially. It usually localizes to the lung apices and lateral aspect of the lungs. Based on this prior clinical knowledge, we propose a template-guided approach that integrates the disease occurrence areas with baseline model explanations. The proposed approach requires minimal human involvement and yields explanations that align better with the clinical understanding of pneumothorax. **Figure 1** shows the overview of our template guidance as a plug-and-play module for existing XAI methods. To depict the pleural space from the clinical experts' view, a canonical lesion annotation by radiologists is extracted as the basis for template generation. Then several morphological operations are implemented to further refine the pleural space - potential occurrence areas of pneumothorax. After that, we shepherd the original explanations using the generated template region: Only the pixel within the template boundaries will be included in model focus areas. Finally, focus areas with or without template guidance are compared with the ground truth lesion annotations.

The first step in the proposed template guidance is to generate the optimal template carrying the location information of disease occurrence. **Figure 2** summarizes the details of template generation: Using one canonical lesion delineation $A_T^{train}$ as the starting point, the candidate templates are generated by flipping, overlap, and dilation. Selected by radiologists, $A_T^{train}$ contours at least the pleural space on one side. Then the step of flipping turns over the original lesion delineation horizontally to generate $A_{T,flip}^{train}$ on the other side. After that, considering the domain knowledge that pneumothorax potentially occurs in both the left and the right pleural space, the step of overlapping is implemented to generate $A_{T,overlap}^{train}$ spotlighting both left and right pleural spaces [28]. A pixel $p_{w,h}(A_{T,overlap}^{train})$ is included in the template area if it is within either $p_{w,h}(A_T^{train})$ or $p_{w,h}(A_{T,flip}^{train})$. Another factor affecting the depiction of pleural space is that the chest radiographs are captured at different distances and angles. Thus, the concrete position and scale of pleural space vary in different radiographs [29]. To address this issue, we introduce the step of dilation to eliminate the problem of deformation through enlarging the template area to cover a broader space. Following the previous work [30], a 15×15 ellipse kernel sweeps each pixel on the original image, and a pixel will be included in the dilated template area if one of its neighbor pixels within the kernel belongs to $A_{T,overlap}^{train}$. After that, we obtain the final template $T^*$ wherein $a_{w,h}(T^*) = 1$ denotes the coordinate of $(w, h)$ in $I_i^{test}$ belongs to the pleural space. Through the element-wise product, the template-guided pixel importance $E^*(p_{w,h}(I_i^{test}))$ is calculated:

$$E^*(p_{w,h}(I_i^{test})) = T^* \odot E(p_{w,h}(I_i^{test})). \qquad (9)$$

Finally, the identical approach as the baseline explanation is employed to extract the model focus region from $E^*(p_{w,h}(I_i^{test}))$.

## 3 Experiments

In this section, we first introduced the datasets. Then we provided details on the training and explanation of pneumothorax classifiers, and clarified the relevant evaluation metrics. After that, we presented the experimental results of pneumothorax classification and explanation. We demonstrated that the proposed template-guided approach consistently improved the baseline XAI methods. To provide a comprehensive assessment, we visualized both successful and collapsing cases of model explanations. All experiments were conducted using Python, and the code has been made publicly available on GitHub for reproducibility [31].

### 3.1 Datasets

The performance of pneumothorax classification and explanation was demonstrated using two real-world datasets of SIIM-ACR Pneumothorax Segmentation Challenge [32] and ChestX-Det [33]. The SIIM-ACR dataset comprises a total of 12,047 chest radiographs, among which 2,669 instances are diagnosed as positive, indicating the presence of pneumothorax. Unlike the SIIM-ACR dataset, The ChestX-Det dataset is notably smaller, consisting of 611 healthy images and 189 pneumothorax-positive images. Besides the binary pneumothorax diagnosis at the image level, both datasets provide pixel-level lesion delineations in positive cases.

We randomly split the SIIM-ACR dataset into three parts at 60: 20: 20. Specifically, the training set consisted of 7,226 images (60%, containing 1,600 positive samples), the validation set comprised 2,410 images (20%, containing 534 positive samples), and the test set included 2,410 images (20%, containing 534 positive samples). To validate the generalizability of the proposed method, we evenly partitioned the ChestX-Det dataset into validation (50%, containing 95 positive samples) and test sets (50%, containing 94 positive samples) for external validation. **Table 1** gives an overview of the used data sets, annotations, and functions in our study. Detailed information is elaborated in the subsequent two sections.

## 3.2 Classifier training and evaluation

We implemented the pneumothorax classifier with two lightweight architectures: VGG-19 [34] and ResNet-50 [15], modifying their outputs for binary classification. A Stochastic Gradient Descent (SGD) optimizer was employed with a learning rate of 1e-3 and a momentum of 0.9. Model training was conducted in batches of 16 images, using weighted cross-entropy as the loss function to counterbalance the predominance of negative samples[35]. The training was set as 100 epochs on the training set of SIIM-ACR with an early-stop initiated if no improvement was observed on the validation set of SIIM-ACR over 10 consecutive epochs. After the training, the model classification performance was evaluated on both the internal test set of SIIM-ACR and the external test set of ChestX-Det. Evaluation metrics included AUROC, the area under the precision recall curve (AUPRC), accuracy, sensitivity, specificity, positive predictive value (PPV), and negative predictive value (NPV). Binarization cutoffs were chosen as the points closest to the upper-left corner in the ROC curves on the respective validation sets [36]. For each metric, standard errors were calculated using the nonparametric bootstrap method [37].

## 3.3 XAI explanation and evaluation

After the pneumothorax classification training and evaluation, model explanations play a pivotal role in gaining the trust of clinicians [22]. Our study utilized three model explanation methods: Saliency Map [20], Grad-CAM [12], and Integrated Gradients [21]. Our template-guided approach worked as a plug-and-play module on the basis of the three XAI methods, necessitating only one lesion delineation from the training set of SIIM-ACR. Therefore, we had a total of six explanation methods. The direct production of the six explanation methods was the pixel importance, from which focus areas were further outlined as the final explanation using the threshold value $v^*$ of 0.95 [38]. We leveraged IoU and Dice Score Coefficient (DSC) to quantify the alignment between the generated focus areas and ground truth lesion delineations on both the internal test set of SIIM-ACR and the external test set of ChestX-Det. The standard errors of IoU and DSC were computed through the nonparametric bootstrap method [37].

## 3.4 Experimental results

In this section, we showed the evaluation results followed by their respective standard errors enclosed within parentheses. **Table 2** quantifies the model classification performance on the internal test set of SIIM-ACR. The VGG-19 classifier achieved results of an AUROC of 0.864 (0.008), an AUPRC of 0.660 (0.023), an accuracy of 80.5% (0.8%), a sensitivity of 78.3% (1.8%), a specificity of 81.1% (0.9%), a PPV of 54.1% (1.9%), and an NPV of 92.9% (0.7%). The ResNet-50 discriminator attained an AUROC of 0.842 (0.007), an AUPRC of 0.630 (0.023), an accuracy of 77.8% (0.8%), a sensitivity of 75.7% (1.5%), a specificity of 78.4% (0.9%), a PPV of 49.9% (2.0%), and an NPV of 91.9% (0.6%). Following the evaluation of model classification, **Table 3** illustrates the model explanation performance of the baseline XAI methods and their template-guided versions. Under the framework of VGG-19, the original Saliency Map achieved an IoU of 2.2% (0.2%) and a DSC of 4.1% (0.3%). The original Grad-CAM obtained an IoU of 1.4% (0.1%) and a DSC of 2.6% (0.2%). The original Integrated Gradients achieved an IoU of 3.1% (0.2%) and a DSC of 5.9% (0.3%). Adding template guidance consistently resulted in performance improvements in terms of IoU and DSC: 1.0% and 1.9% for Saliency Map, 0.9% and 1.7% for Grad-CAM, and 1.4% and 2.3% for Integrated Gradients. Based on ResNet-50, the performance enhancements were 1.7% and 3.1% for Saliency Map, 3.0% and 5.1% for Grad-CAM, and 2.6% and 4.5% for Integrated Gradients. In the internal test scenarios, the incremental percentages of IoU and DSC, calculated by the performance improvements over the baseline performance, ranged from 41.7% to 168.4% and 30.7% to 114.9%, respectively.

**Table 4** presents the classification performance of developed VGG-19 and ResNet-50 on the external test set of ChestX-Det. Specifically, VGG-19 without fine-tuning presented an AUROC of 0.942 (0.016), an AUPRC of 0.896 (0.025), an accuracy of 89.7% (1.5%), a sensitivity of 86.2% (3.3%), a specificity of 90.8% (1.6%), a PPV of 74.3% (4.4%), and an NPV of 95.5% (1.1%). The directly-deployed ResNet-50 also showed satisfactory performance. Regarding the explanation performance, akin to the internal validation on SIIM-ACR, our template-guided approach consistently improved all three baseline XAI methods as showcased in **Table 5**. In terms of IoU and DSC, the template-guided explanation of VGG-19 achieved improvements of 1.6% and 3.0% for Saliency Map, 0.8% and 1.7% for Grad-CAM, and 1.6% and 2.9% for Integrated Gradients. Based on ResNet-50, the performance enhancements were 2.0% and 3.6% for Saliency Map, 2.0% and 3.6% for Grad-CAM, and 2.4% and 4.2% for Integrated Gradients. Notably, the incremental percentages of IoU and DSC, when compared with baseline methods, ranged from 71.3% to 130.9% and 66.5% to 134.1%, respectively. We highlighted that the used template was the one depicted in **Figure 2** while other template candidates, as shown in the Appendix, also yielded comparable improvements, presenting the robustness of the proposed approach.

These quantitative metrics elucidated the explanation improvements attributable to the proposed template-guided approach. To further compare XAI methods with and without template guidance, **Figure 3** and **Figure 4** visualize their explanations on the internal test set of SIIM-ACR and the external test set of ChestX-Det, respectively. From the left to the right, each figure displays the lesion areas delineated by radiologists (Ground truth), important regions outlined by the original explanations (Baseline), and the enhanced explanations (Our method). **Figure 3b** and **Figure 4b** show the samples on which the proposed approach can upgrade the original explanation quality. However, the proposed method fails to upgrade the baseline in **Figure 3c** and **Figure 4c**. Such a performance contrast demonstrated that our template-guided approach collapses when the pneumothorax exists outside the template region. **Figure 3a** and **Figure 4a** illustrate scenarios where both XAI methods with and without template guidance perform well, whereas **Figure 3d** and **Figure 4d** depict situations where both XAI methods with and without template guidance exhibit poor performance. Also, we identified that either method presented a lower performance for small pneumothorax compared with the large one, which has been reported by other studies [2].

## 4 Discussion

This study proposed a template-guided approach to improve AI explanations in the context of pneumothorax diagnoses. Based on clinical knowledge that pneumothorax occurs in the pleural space, we generated a template covering the pleural space based on a canonical lesion annotation by radiologists. Then the template was superimposed on the baseline explanations to filter out extraneous model explanations that fall outside the template's boundaries. This straightforward approach effectively constrained model explanations within the potential areas of pneumothorax occurrence, thereby consistently improving baseline XAI methods across twelve benchmark scenarios.

Beyond the investigated pneumothorax, our template guidance holds applicability for other thoracic diseases characterized by clinically validated disease locations. Cardiomegaly, the heart enlargement evident at the cardiac region [39], serves as another use case for the proposed template-guided approach. According to the radiological knowledge, the cardiac region encompasses the central area of a frontal chest radiograph [40]. To derive a comprehensive occurence template of cardiomegaly, radiologists are invited to meticulously analyze radiograph samples and collaborate closely with ML engineers to ascertain the details of morphological operations. With the derived template, model explanation aligns better with clinical knowledge, underscoring the value of embedding domain knowledge in DL for healthcare.

Our method focuses on incorporating domain knowledge into model explanations. Nevertheless, it is worth noting that some professionals hold the perspective that a segmentation model assists clinicians better than a classification model supplemented with explanations [41]. Unlike a classification model that outputs a single diagnostic probability, a segmentation model explicitly delineates disease lesion areas. Yet an accurate segmentation model is largely dependent on the availability of large-scale pixel-level annotations, which are time-consuming and hard to acquire [42]. Potential solutions to this dilemma are semi-supervised learning and weakly-supervised learning [43].

Validated through comprehensive studies, both semi-supervised learning and weakly-supervised learning stand out as effective methods for alleviating the annotation burden during the development of an accurate segmentation model. Madani et al. [41] proposed a semi-supervised approach for cardiac disease prediction that achieved high accuracy using only a small amount of lesion annotations. Based on only 4% labeled data, they achieved 85% of the accuracy by the fully-supervised model on 100% labeled data. Semi-supervised learning still requires few pixel-level annotations while weakly-supervised learning aims to build a segmentation model using only image-level labels. Ouyang et al. [44] derived the pixel-level segmentations through focus areas extracted from a classification model and corrected the noisy segmentation label by a spatial annotation smoothing technique. They showed that the weakly-supervised approach upgraded the segmentation model training significantly without any pixel-level annotations. While these methods are promising in reducing the labeling cost, several studies have reported that semi or weakly-supervised learning failed to reach the baseline by a fully-supervised model [45,46]. In medical artificial intelligence, how to achieve a balance between the annotation cost and AI accuracy remains an unsolved conundrum in resource-limited settings [25]. With the recent release of versatile foundation models, a potential solution could be the Segment Anything Model (SAM), known for its capability to cut out any object in any image with a single click [47]. Hence, under the same budget, SAM facilitates the annotation of a larger number of samples and the development of a more accurate segmentation model [47].

Our study has limitations that warrant future investigation. First, we employed a static template as a prior in guiding model explanations. Although the proposed method improved baseline explanations, the current performance is still unable to meet the deployment standards required by some regulatory agencies. For example, the Korea Ministry of Food and Drug Safety mandates a minimum Dice coefficient of 20% for clinically relevant explanations [48]. Future research may explore the integration of the current approach with affine transformation, which has been proven valuable in modifying the scale, angle, and displacement of the fixed template, thereby improving the explanation performance [18,29,49]. Second, we evaluated the performance of the template-guided approach within a limited set of experimental configurations, comprising three XAI baselines, two DL models, and one thoracic disease. Future endeavors will encompass alternative XAI methods like LayerCAM [50], extra DL models including vision transformer [51], and additional thoracic diseases such as cardiomegaly [52-54] for a more comprehensive assessment of our approach.

## 5 Conclusion

Historically, clinical domain knowledge was undervalued by the DL community when designing XAI methods for DL-aided diagnoses. In this study, we showcase the value of clinical knowledge, especially potential areas of disease occurrence, in consistently improving model explanations across twelve benchmark scenarios. It is imperative to highlight that our template-guided approach necessitates only a single lesion delineation crafted by radiologists, obviating the need for extensive annotation efforts. We anticipate that our template guidance will forge a fresh approach to elucidating AI models by integrating clinical domain expertise.

**Funding**

This study was supported by Duke-NUS Medical School, Singapore. The funder had no role in study design, data collection and analysis, decision to publish, or preparation of the manuscript.

**CRediT authorship contribution statement**

**Han Yuan**: Conceptualization, Data curation, Formal analysis, Investigation, Methodology, Software, Validation, Visualization, Writing - original draft, Writing - review editing. **Chuan Hong**: Investigation, Methodology, Writing - original draft, Writing - review editing. **Pengtao Jiang**: Investigation, Methodology, Writing - review editing. **Gangming Zhao**: Investigation, Methodology, Writing - review editing. **Nguyen Tuan Anh Tran**: Investigation, Methodology, Writing - review editing. Xinxing Xu: Investigation, Methodology, Writing - review editing. **Yet Yen Yan**: Investigation, Methodology, Writing - review editing. **Nan Liu**: Conceptualization, Formal analysis, Funding acquisition, Investigation, Methodology, Project administration, Resources, Supervision, Validation, Writing - original draft, Writing - review editing.

**Declaration of competing interest**

All authors declare that they have no conflicts of interest.

**Declaration of generative AI in scientific writing**

During the revision of the initial draft, Han Yuan used GPT-3.5 to check grammar. After using this tool, Han Yuan and other authors reviewed and edited the content as needed. Han Yuan takes full responsibility for the content of the publication.

Table 1: An overview of the used data set, annotation, and function

| Data set | | Annotation | Function |
|---|---|---|---|
| SIIM-ACR | Training set | Binary diagnosis | Classifier training |
| | | Lesion delineation | Template generation |
| | Validation set | Binary diagnosis | Classifier training |
| | | | Binarization cutoff calculation |
| | Test set | Binary diagnosis | Internal evaluation of classifier's classification capability |
| | | Lesion delineation | Internal evaluation of XAI's explanation capability |
| ChestX-Det | Validation set | Binary diagnosis | Binarization cutoff calculation |
| | Test set | Binary diagnosis | External evaluation of classifier's classification capability |
| | | Lesion delineation | External evaluation of XAI's explanation capability |

Table 2: Internal classification evaluation of various deep learning models. The evaluation metrics on the test set are presented, accompanied by their respective standard errors enclosed within parentheses.

| Model | AUROC | AUPRC | Accuracy (%) | Sensitivity (%) | Specificity (%) | PPV (%) | NPV (%) |
|---|---|---|---|---|---|---|---|
| VGG-19 | 0.864 (0.008) | 0.660 (0.023) | 80.5 (0.8) | 78.3 (1.8) | 81.1 (0.9) | 54.1 (1.9) | 92.9 (0.7) |
| ResNet-50 | 0.842 (0.007) | 0.630 (0.023) | 77.8 (0.8) | 75.7 (1.5) | 78.4 (0.9) | 49.9 (2.0) | 91.9 (0.6) |

Table 3: Internal explanation evaluation of various deep learning models by XAI methods. The evaluation metrics on the test set are presented, accompanied by their respective standard errors enclosed within parentheses.

| Model | XAI | Knowledge-Guidance | IoU (%) | DSC (%) |
|---|---|---|---|---|
| VGG-19 | Saliency Map | × | 2.2 (0.2) | 4.1 (0.3) |
| | | √ | 3.2 (0.2) | 6.0 (0.3) |
| | Grad-CAM | × | 1.4 (0.1) | 2.6 (0.2) |
| | | √ | 2.3 (0.2) | 4.3 (0.3) |
| | Integrated Gradients | × | 3.1 (0.2) | 5.9 (0.3) |
| | | √ | 4.5 (0.2) | 8.2 (0.3) |
| ResNet-50 | Saliency Map | × | 2.3 (0.1) | 4.3 (0.2) |
| | | √ | 4.0 (0.2) | 7.4 (0.3) |
| | Grad-CAM | × | 1.7 (0.2) | 3.1 (0.3) |
| | | √ | 4.7 (0.3) | 8.2 (0.4) |
| | Integrated Gradients | × | 2.1 (0.1) | 4.0 (0.2) |
| | | √ | 4.7 (0.2) | 8.5 (0.4) |

Table 4: External classification evaluation of various deep learning models. The evaluation metrics on the test set are presented, accompanied by their respective standard errors enclosed within parentheses.

| Model | AUROC | AUPRC | Accuracy (%) | Sensitivity (%) | Specificity (%) | PPV (%) | NPV (%) |
|-------|-------|-------|--------------|-----------------|-----------------|---------|---------|
| VGG-19 | 0.942 (0.016) | 0.896 (0.025) | 89.7 (1.5) | 86.2 (3.3) | 90.8 (1.6) | 74.3 (4.4) | 95.5 (1.1) |
| ResNet-50 | 0.943 (0.013) | 0.870 (0.029) | 89.7 (1.6) | 84.0 (3.8) | 91.5 (1.7) | 75.2 (3.9) | 94.9 (1.2) |

Table 5: External explanation evaluation of various deep learning models by XAI methods. The evaluation metrics on the test set are presented, accompanied by their respective standard errors enclosed within parentheses.

| Model | XAI | Knowledge-Guidance | IoU (%) | DSC (%) |
|-------|-----|--------------------|---------|---------|
| VGG-19 | Saliency Map | × | 1.3 (0.2) | 2.5 (0.4) |
| | | √ | 2.9 (0.3) | 5.5 (0.6) |
| | Grad-CAM | × | 1.1 (0.4) | 1.9 (0.4) |
| | | √ | 1.9 (0.4) | 3.6 (0.7) |
| | Integrated Gradients | × | 2.3 (0.3) | 4.4 (0.5) |
| | | √ | 3.9 (0.4) | 7.3 (0.7) |
| ResNet-50 | Saliency Map | × | 1.7 (0.2) | 3.2 (0.4) |
| | | √ | 3.7 (0.4) | 6.8 (0.7) |
| | Grad-CAM | × | 1.5 (0.4) | 2.7 (0.8) |
| | | √ | 3.5 (0.6) | 6.3 (0.9) |
| | Integrated Gradients | × | 1.8 (0.2) | 3.4 (0.4) |
| | | √ | 4.2 (0.5) | 7.6 (0.8) |

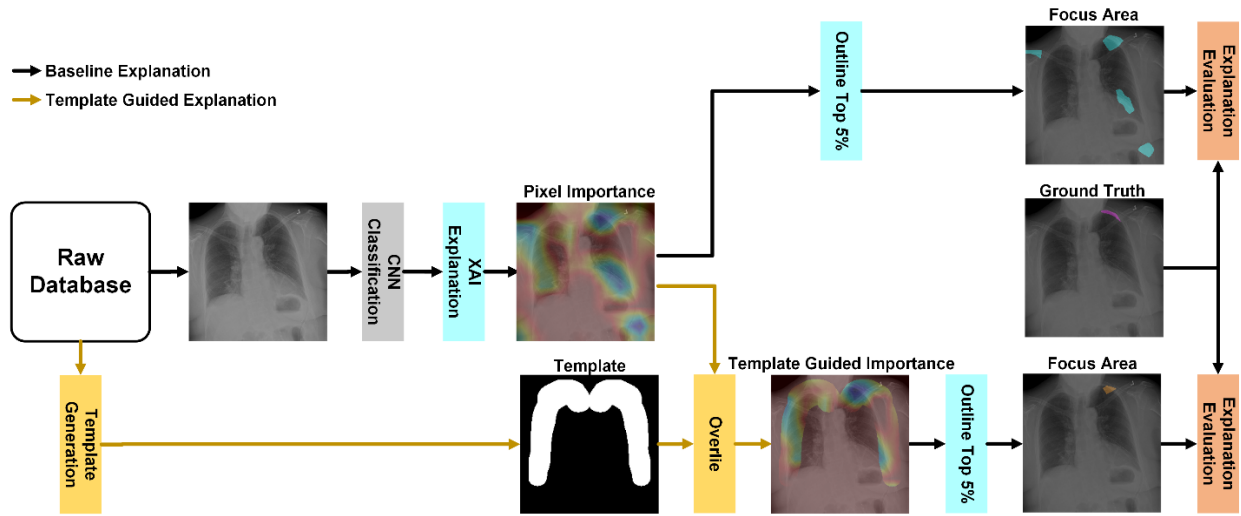

Figure 1: Overview of the proposed template-guided explanation pipeline.

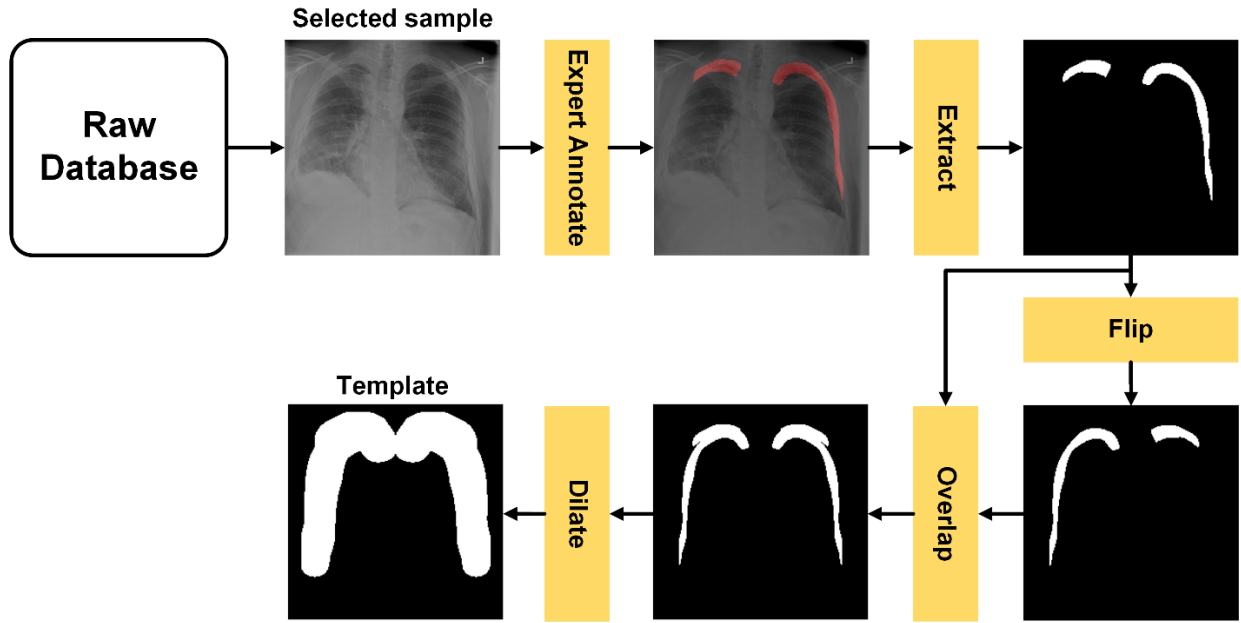

Figure 2: Detailed steps of the template generation.

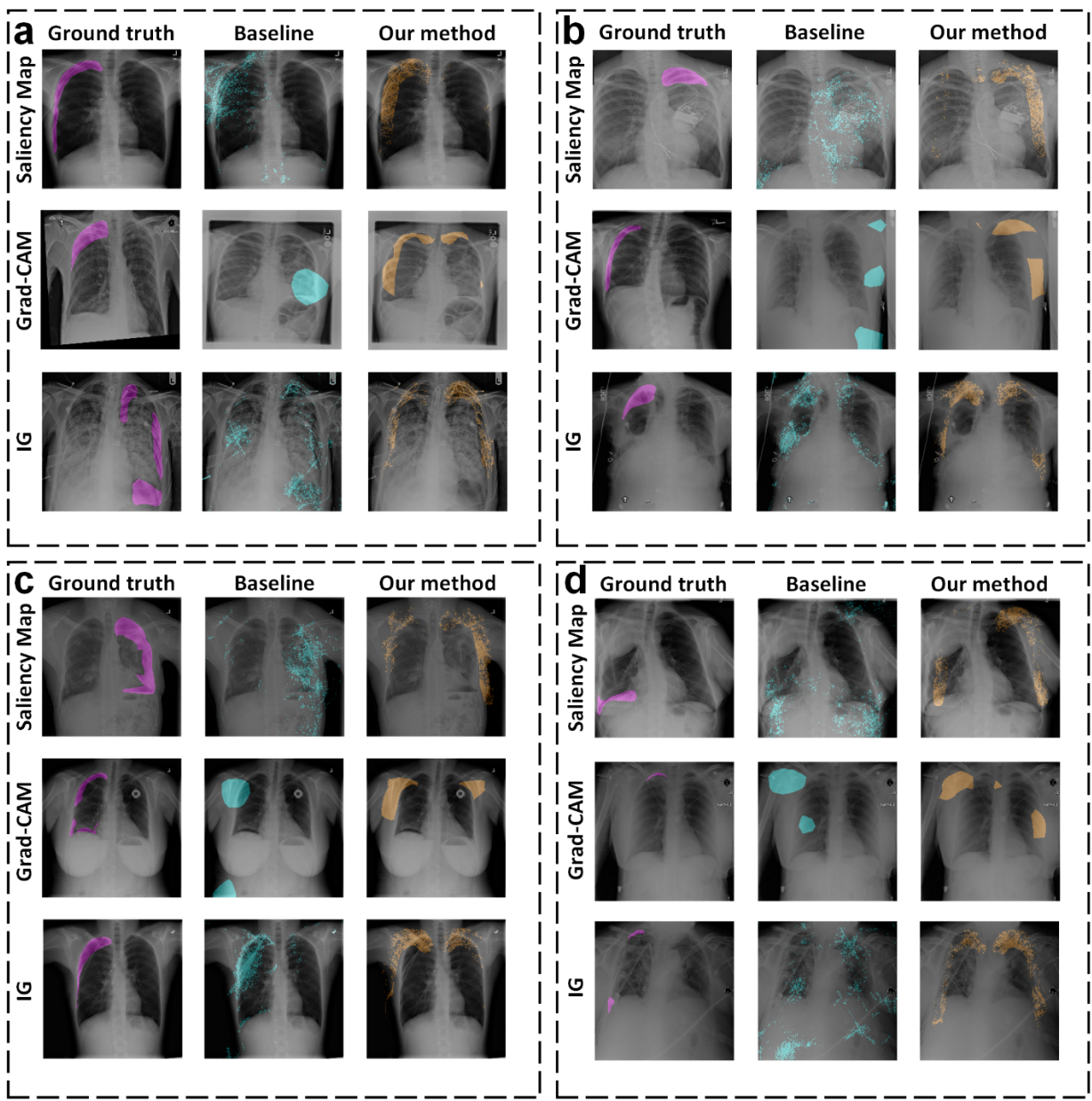

Figure 3: Visualization comparison of pneumothorax radiograph explained by original and the template-guided Saliency Map, Grad-CAM, and Integrated Gradients (IG) on the internal test set of SIIM-ACR.

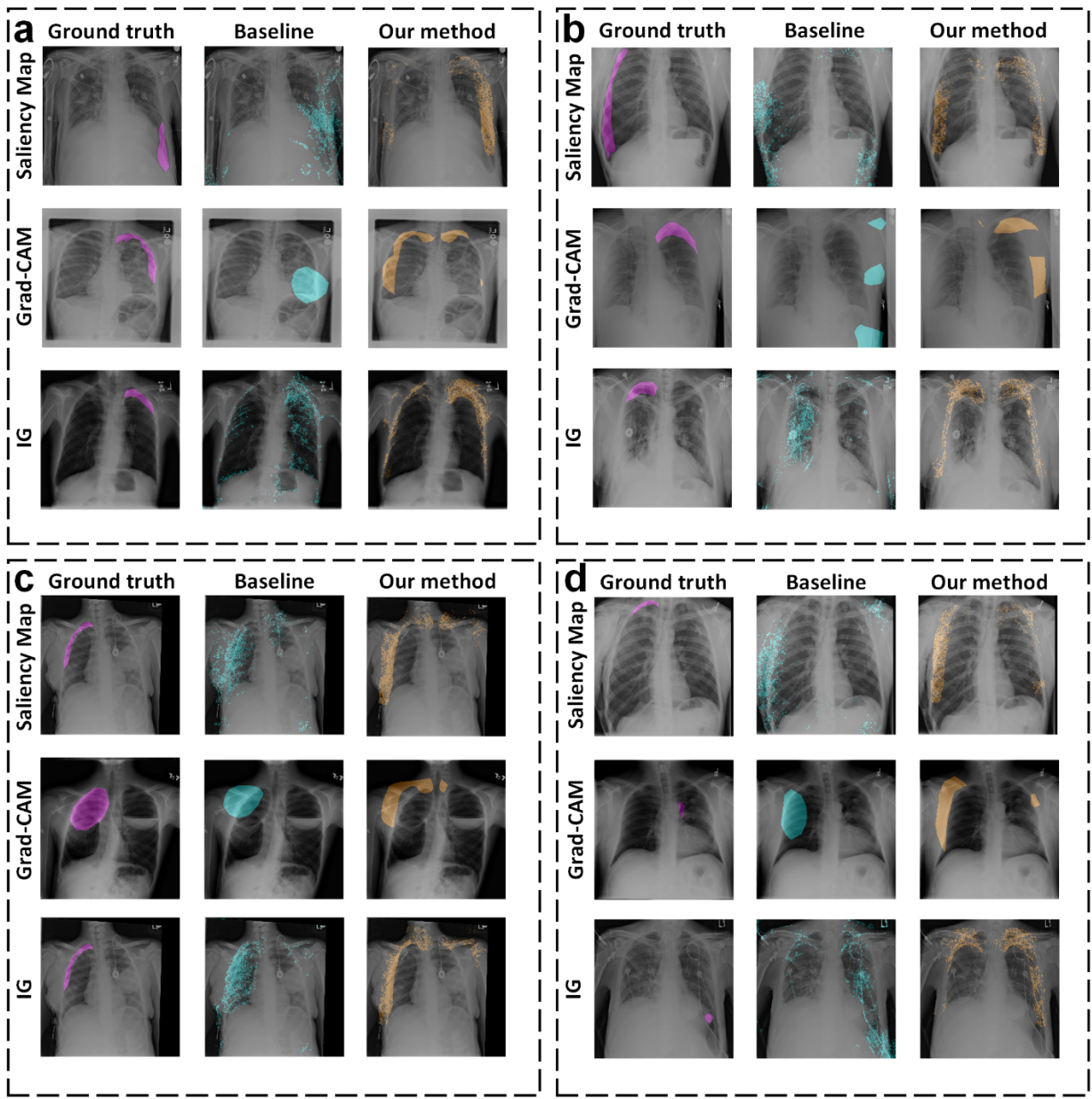

Figure 4: Visualization comparison of pneumothorax radiograph explained by original and the template-guided Saliency Map, Grad-CAM, and Integrated Gradients (IG) on the external test set of ChestX-Det.

**Appendix**

Figure A.1 presents four alternative guidance templates and their performance in upgrading model explanation quality on SIIM-ACR and ChestX-Det datasets are shown in Table A.1 and A.2, respectively.

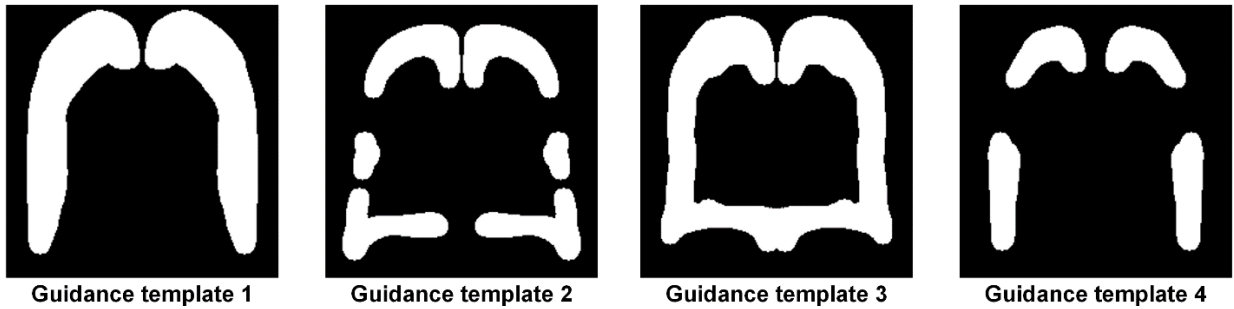

Figure A.1 Alternative examples of guidance templates

Table A.1 Internal explanation evaluation of various deep learning models by XAI methods. The evaluation metrics on the test set are presented, accompanied by their respective standard errors enclosed within parentheses.

| Guidance Index | XAI | Knowledge-Guidance | IoU (%) | DSC (%) |
|---|---|---|---|---|
| VGG-19 | Saliency Map | × | 2.22 (0.15) | 4.14 (0.26) |
| | | Template 1 | 2.94 (0.15) | 5.46 (0.26) |
| | | Template 2 | 2.73 (0.13) | 5.11 (0.26) |
| | | Template 3 | 2.84 (0.15) | 5.30 (0.26) |
| | | Template 4 | 3.57 (0.15) | 6.60 (0.26) |
| | Grad-CAM | × | 1.43 (0.13) | 2.63 (0.20) |
| | | Template 1 | 2.24 (0.20) | 4.07 (0.33) |
| | | Template 2 | 2.32 (0.18) | 4.25 (0.31) |
| | | Template 3 | 2.27 (0.18) | 4.14 (0.33) |
| | | Template 4 | 2.70 (0.23) | 4.91 (0.36) |
| | Integrated Gradients | × | 3.14 (0.18) | 5.85 (0.31) |
| | | Template 1 | 4.53 (0.18) | 8.34 (0.31) |
| | | Template 2 | 4.27 (0.15) | 7.89 (0.26) |
| | | Template 3 | 4.52 (0.18) | 8.34 (0.28) |
| | | Template 4 | 4.96 (0.18) | 9.08 (0.28) |
| ResNet-50 | Saliency Map | × | 2.30 (0.10) | 4.33 (0.20) |
| | | Template 1 | 3.99 (0.18) | 7.33 (0.31) |
| | | Template 2 | 3.68 (0.15) | 6.81 (0.26) |
| | | Template 3 | 3.43 (0.15) | 6.36 (0.26) |
| | | Template 4 | 4.98 (0.20) | 9.01 (0.36) |
| | Grad-CAM | × | 1.74 (0.15) | 3.09 (0.28) |
| | | Template 1 | 4.71 (0.31) | 8.12 (0.48) |
| | | Template 2 | 4.00 (0.26) | 7.08 (0.41) |
| | | Template 3 | 4.08 (0.28) | 7.14 (0.43) |
| | | Template 4 | 5.33 (0.28) | 9.29 (0.46) |
| | Integrated Gradients | × | 2.09 (0.13) | 3.95 (0.23) |
| | | Template 1 | 4.70 (0.20) | 8.56 (0.36) |
| | | Template 2 | 4.54 (0.18) | 8.31 (0.31) |
| | | Template 3 | 4.50 (0.18) | 8.25 (0.33) |
| | | Template 4 | 5.46 (0.20) | 9.84 (0.36) |

Table A.2 External explanation evaluation of various deep learning models by XAI methods.

| Guidance Index | XAI | Knowledge-Guidance | IoU (%) | DSC (%) |
|---|---|---|---|---|
| VGG-19 | Saliency Map | × | 1.33 (0.20) | 2.54 (0.36) |
| | | Template 1 | 2.46 (0.36) | 4.61 (0.61) |
| | | Template 2 | 2.36 (0.33) | 4.42 (0.59) |
| | | Template 3 | 2.34 (0.31) | 4.43 (0.54) |
| | | Template 4 | 2.86 (0.38) | 5.36 (0.64) |
| | Grad-CAM | × | 1.05 (0.36) | 1.88 (0.59) |
| | | Template 1 | 1.73 (0.33) | 3.21 (0.61) |
| | | Template 2 | 1.79 (0.36) | 3.33 (0.64) |
| | | Template 3 | 1.66 (0.33) | 3.09 (0.61) |
| | | Template 4 | 2.15 (0.38) | 3.97 (0.66) |
| | Integrated Gradients | × | 2.30 (0.26) | 4.39 (0.46) |
| | | Template 1 | 3.94 (0.38) | 7.35 (0.69) |
| | | Template 2 | 3.91 (0.36) | 7.23 (0.61) |
| | | Template 3 | 3.87 (0.36) | 7.21 (0.64) |
| | | Template 4 | 4.18 (0.38) | 7.75 (0.64) |
| ResNet-50 | Saliency Map | × | 1.67 (0.23) | 3.18 (0.41) |
| | | Template 1 | 3.54 (0.38) | 6.56 (0.66) |
| | | Template 2 | 3.25 (0.38) | 6.06 (0.69) |
| | | Template 3 | 2.92 (0.31) | 5.50 (0.56) |
| | | Template 4 | 4.39 (0.46) | 8.05 (0.77) |
| | Grad-CAM | × | 1.52 (0.43) | 2.67 (0.77) |
| | | Template 1 | 3.77 (0.59) | 6.66 (0.97) |
| | | Template 2 | 3.39 (0.66) | 5.91 (1.05) |
| | | Template 3 | 3.51 (0.64) | 6.16 (1.05) |
| | | Template 4 | 4.54 (0.77) | 7.88 (1.22) |
| | Integrated Gradients | × | 1.80 (0.23) | 3.41 (0.43) |
| | | Template 1 | 4.27 (0.51) | 7.81 (0.87) |
| | | Template 2 | 4.17 (0.48) | 7.60 (0.84) |
| | | Template 3 | 4.03 (0.51) | 7.39 (0.89) |
| | | Template 4 | 4.78 (0.56) | 8.68 (0.94) |